%% file: main.tex
\documentclass{article}

\usepackage[final,nonatbib]{neurips_2024}

\usepackage[utf8]{inputenc} 
\usepackage[T1]{fontenc}    
\usepackage{hyperref}       
\usepackage{url}            
\usepackage{booktabs}       
\usepackage{amsfonts}       
\usepackage{nicefrac}       
\usepackage{microtype}      
\usepackage{xcolor}         
\usepackage[pdftex]{graphicx}
\usepackage{amsmath} 
\usepackage{adjustbox} 
\usepackage{cleveref}       
\usepackage{wrapfig}

\include{custom_def}

\title{Learning Transferable Features for \\ Implicit Neural Representations}

\author{%
  Kushal Vyas \\
  \texttt{kushal.vyas@rice.edu} \\
  \And
  Ahmed Imtiaz Humayun \\
  \texttt{imtiaz@rice.edu} \\
  \And
  Aniket Dashpute \\
  \texttt{aniket.dashpute@rice.edu} \\
  \AND
  Richard G. Baraniuk \\
  \texttt{richb@rice.edu} \\
  \And
  Ashok Veeraraghavan \\
  \texttt{vashok@rice.edu} \\
  \And
  Guha Balakrishnan\\
  \texttt{guha@rice.edu}\\
  \AND
  {}\\
  \textbf{Rice University}
}

\begin{document}

\maketitle

\begin{abstract}
 Implicit neural representations (INRs) have demonstrated success in a variety of applications, including inverse problems and neural rendering. An INR is typically trained to capture one signal of interest, resulting in learned neural features that are highly attuned to that signal. Assumed to be less \emph{generalizable}, we explore the aspect of transferability of such learned neural features for fitting similar signals. We introduce a new INR training framework, \strainer that learns transferrable features for fitting INRs to new signals from a given distribution, faster and with better reconstruction quality. Owing to the sequential layer-wise affine operations in an INR, we propose to learn transferable representations by sharing initial \textit{encoder} layers across multiple INRs with independent \textit{decoder} layers. At test time, the learned \textit{encoder} representations are transferred as initialization for an otherwise randomly initialized INR. We find \strainer to yield extremely powerful initialization for fitting images from the same domain and allow for a $\approx +10dB$ gain in signal quality early on compared to an untrained INR itself. \strainer also provides a simple way to encode data-driven priors in INRs. We evaluate \strainer on multiple in-domain and out-of-domain signal fitting tasks and inverse problems and further provide detailed analysis and discussion on the transferability of \strainers features. Our demo can be accessed \href{https://kushalvyas.github.io/strainer.html}{here.}
 \end{abstract}

\section{Introduction} \label{sec:Intro}
\input{content/1_Introduction/intro}

\section{Background} \label{sec:Related_Work}
\input{content/2_RelatedWorks/relatedworksv2}


\section{Methods} \label{Method}
\input{content/3_Method/method}

\section{Experiments} \label{sec:ExperimentsResults}
\input{content/4_ExperimentResults/Results}

\section{Discussion and Conclusion} \label{sec:Discussions}
\input{content/7_Discussions/Discussions}

\section{Broader Impacts}\label{sec:BroaderImpact}
\input{content/6_BroaderImpact/BroaderImpact}

\section*{Acknowledgments and Disclosure of Funding}
\input{content/Acknowledgements/Ack}

\bibliography{references/references}

\include{content/Supplementary/supp}
\newpage

\end{document}

%% file: custom_def.tex
\newcommand{\siren}{{\sc siren\ }}
\newcommand{\sirenB}{{\sc \textbf{siren}\ }}
\newcommand{\sirenn}{{\sc siren}}
\newcommand{\citesiren}{\siren\cite{sitzmann2020implicit}\ }
\newcommand{\strainer}{{\sc strainer\ }}
\newcommand{\strainerB}{{\sc \textbf{strainer}\ }}
\newcommand{\strainerBs}{{\sc \textbf{strainer's}\ }}
\newcommand{\strainers}{{\sc strainer}'s\ }
\newcommand{\strainerten}{{\sc strainer-10}\ }
\newcommand{\celeba}{CelebA-HQ\ }
\newcommand{\metalearning}{Meta-learned 5K\ }
\newcommand{\TESTCASES}{550\ }
\newcommand{\TESTCASESAFHQ}{368\ }
\newcommand{\TESTCASESMRI}{144 \ }
\newcommand{\encoderimages}{10 \ }

\newcommand{\transinrdefault}{Trans INR w/o TTO}
\newcommand{\transinr}{Trans INR w TTO}
\newcommand{\ipcreludefault}{IPC(ReLU + Pos Enc.) w/o TTO}
\newcommand{\ipcrelu}{IPC(ReLU + Pos Enc.) w TTO}

%% file: content/1_Introduction/intro.tex
Implicit neural representations (INRs) are a powerful family of continuous learned function approximators for signal data that are implemented using multilayer perceptron (MLP) deep neural networks. An INR $f_\theta: \mathbb{R}^{m} \mapsto \mathbb{R}^{n}$ maps \emph{coordinates} lying in a $m$-dimensional space to a value in a $n$-dimensional output space, where $\theta$ represents the MLP's tunable parameters. For example, a typical INR for a natural image would use an input space in $\mathbb{R}^{2}$ (consisting of the $x$ and $y$ pixel coordinates), and an output space in $\mathbb{R}^{3}$ (representing the RGB value of a pixel). 
INRs have demonstrated several useful properties including capturing details at all spatial frequencies~\cite{sitzmann2020implicit, saragadam2023wire}, providing powerful priors for natural signals~\cite{saragadam2023wire,sitzmann2020implicit}, and facilitating compression~\cite{dupont2021coin, maiya2023nirvana}. For these reasons, in the past 5 years, INRs have found important uses in image and signal processing including shape representation~\cite{genova2019learning, genova2020local}, novel view synthesis~\cite{mildenhall2020nerf,peng2020convolutional,srinivasan2021nerv}, material rendering~\cite{kuznetsov2021neumip}, computational imaging~\cite{attal2021torf,mildenhall2022nerf}, medical imaging~\cite{wang2022neural}, linear inverse problems~\cite{chen2021learning,sun2021coil}, virtual reality~\cite{deng2022fov} and compression~\cite{dupont2021coin,maiya2023nirvana,strumpler2022implicit,zhang2022implicit}.

A key difference between training INRs compared to other neural architectures like CNNs and Transformers is that a single INR is trained on a single signal. The features learned by an INR, therefore, are finely tuned to the morphology of just the one signal it represents. SplineCAM~\cite{humayun2023splinecam} shows that INRs learn to finely partition the input coordinate space by essentially overfitting to the spatial gradients (edges) of the signal. 
While this allows an INR to represent its signal with high fidelity, its features can not ``transfer'' in any way to represent a second signal, even with similar content. If INRs could exhibit elements of transfer learning, as is the case with CNNs and Transformers, their potential would dramatically increase, such as by encoding data distribution priors for inverse imaging problems.

\begin{figure}
    \centering
    \includegraphics[width=\linewidth]{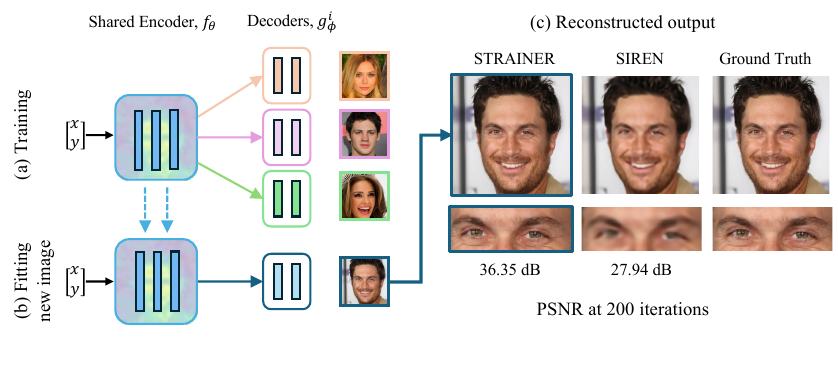}
    \caption{\strainerB - \textbf{Learning Transferable Features for Implicit Neural Representations.} During training time (a), \strainer divides an INR into \textit{encoder} and \textit{decoder layers}. \strainer fits similar signals while sharing the encoder layers, capturing a rich set of transferrable features. At test-time, \strainer serves as powerful initialization for fitting a new signal (b). An INR initialized with \strainers learned encoder features achieves (c) faster convergence and better quality reconstruction compared to baseline \siren models.}
    \label{fig:coverfigure}
    \vspace{-5pt}
\end{figure}

In this work, we take a closer look at INRs and transferable features, and demonstrate that the \emph{first several layers} of an INR can be readily transferred from one signal to another from a domain when trained in a shared setting. We propose \strainer, a simple INR training framework to do so (see \Cref{fig:coverfigure}). \strainer separates an INR into two parts: an ``encoder'' that maps coordinates to features, and a ``decoder'' that maps those features to output values. We fit the encoder over a number of training signals ($1$ to $10$ in our experiments) from a domain, e.g., face images, with separate decoders for each signal. At test time, we initialize a new INR for the test signal consisting of the trained encoder and a randomly initialized decoder. This INR may then be further optimized according to the application of interest. \strainer{} offers a simple and general means of encoding data-driven priors into an INR's parameter initialization.

We empirically evaluate \strainer in several ways. First, we test \strainer on image fitting across several datasets including faces (CelebA-HQ) and medical images (OASIS-MRI) and show (\Cref{fig:psnr_plot_all}) that \strainers learned features are indeed transferrable resulting in a $\approx$+10dB gain in reconstruction quality compared to a vanilla \siren model . We further assess the data-driven prior captured by \strainer by evaluating it on inverse problems such as denoising and super resolution. Lastly, we provide a detailed exploration of how \strainer learns transferable features by exploring INR training dynamics. We conclude by discussing consequences of our results for the new area of INR feature transfer. 

%% file: content/2_RelatedWorks/relatedworksv2.tex
\textbf{Implicit neural representations. } We define $f_\theta(p)$ as an implicit neural representation (INR) \cite{mildenhall2020nerf, sitzmann2020implicit, mescheder2019occupancy} where $f_{\theta}$ is a multi-layer perceptron (MLP) with randomly initialized weights $\theta$ and $p$ is the $m$-dimensional coordinates for the signal.  Each layer in the MLP is an affine operation followed by a nonlinearity such as ReLU \cite{mildenhall2020nerf}, or sine \cite{sitzmann2020implicit}. Given an $n$-dimensional signal $I(p)$, the INR learns a mapping $f: \mathcal{R}^m \rightarrow \mathcal{R}^n$. The INR is iteratively trained to fit the signal by minimizing a loss function such as  ${L}_2$ loss between the signal $I(p)$ and its estimate $f_\theta(p)$:
\begin{equation}
    \theta^* = \arg\min_{\theta} \sum_{i} \mid\mid \mathcal{A}f_\theta(p_i)  - I(p_i) \mid\mid^2_2 \text{ ,}
    \label{eq:L2_loss}
\end{equation}
where $p_i$'s $\in \mathcal{R}^m$ span the given coordinates, $\theta^*$ are the optimal weights that represent the signal, and $\mathcal{A}$ is a differentiable forward model operator such as identity for signal representation and a downsampling operation for inverse problems such as super-resolution. 


\begin{wrapfigure}{R}{0.5\textwidth}
  \centering
    \includegraphics[width=0.5\textwidth]{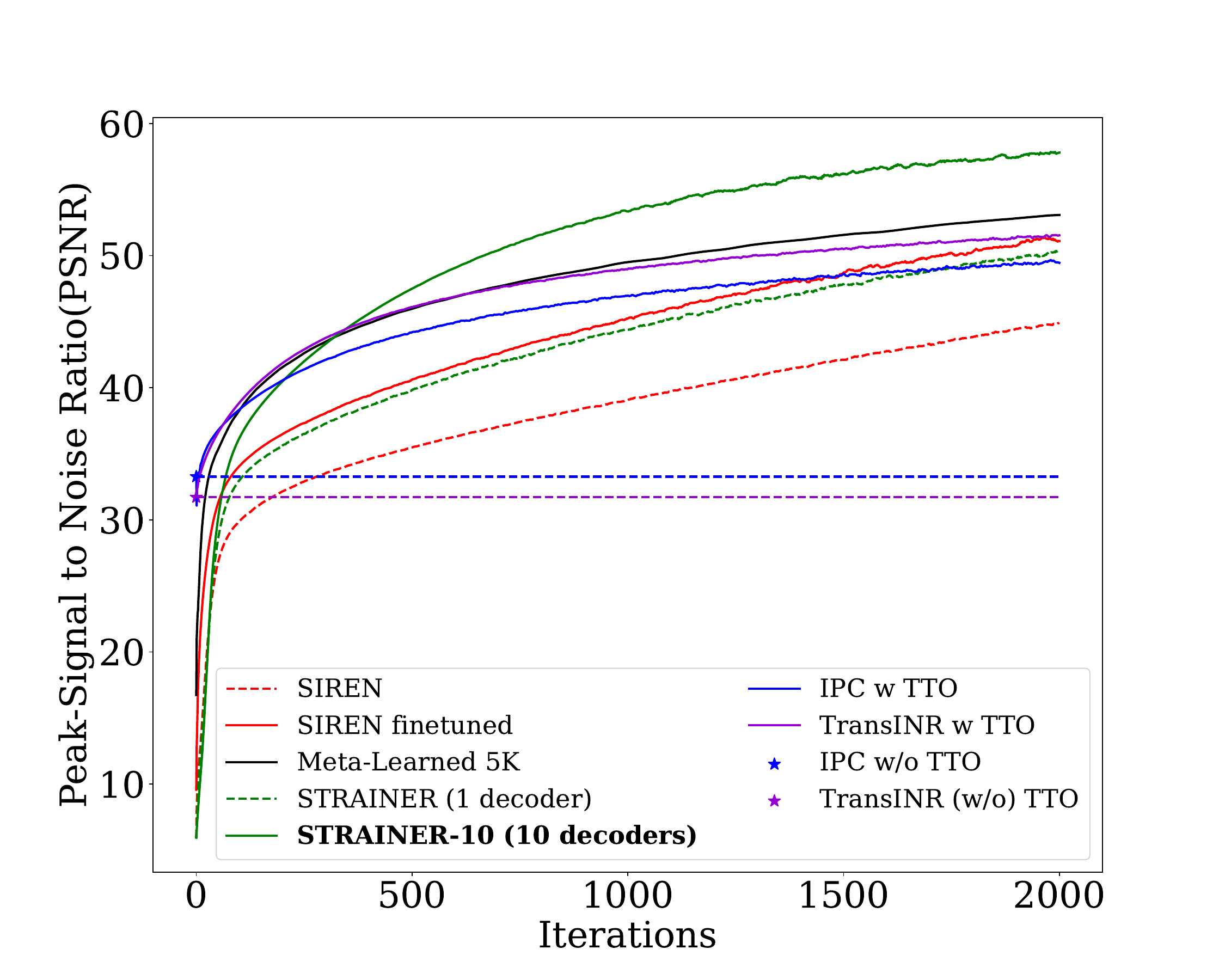}
    
  \caption{\textbf{\strainerB learns faster.} We show the reconstruction quality (PSNR) of different initialization schemes for in-domain image fitting on \celeba\cite{celebahq}. We compare \citesiren  model initialized by (1) random weights (\sirenn), (2) fitting on another face image (\siren finetuned), (3) \strainer-1 (trained using one face image), and (4) \strainerten (trained using ten face images). We also evaluate against multiple baselines such as Meta-Learned 5K \cite{tancik2021learned}, TransINR\cite{transinr}, and IPC\cite{ipc}}
  \label{fig:psnr_plot_all}
  \vspace{-10pt}
\end{wrapfigure}

\textbf{Representation capacity of INRs. }The representation capacity of an INR can be described as the quality of signal the INR can represent within some number of iterations. ReLU-based INRs suffer from spectral bias during training\cite{liu2023finer}, preferentially learning low frequency details in a signal and thus leading to a blurry reconstruction of the represented signal. Fourier positional encodings~\cite{jacot2018neural,mildenhall2020nerf, tancik2020fourfeat} or sinusoidal activation functions (\siren)~\cite{sitzmann2020implicit} help better capture high frequency information. 

Recent works increase the representation capacity of INRs with activations flexible in the frequency domain. WIRE~\cite{saragadam2023wire} uses a continuous Gabor wavelet-based nonlinearity, and demonstrates impressive results for a range of forward and inverse problems. FINER \cite{liu2023finer} splits the sine nonlinearity to have a flexible frequency coverage, and DINER\cite{xie2023diner} uses a hash map to non-uniformly map the input coordinates to a feature vector, effectively re-arranging the spatial arrangement of frequencies and leading to faster and better reconstruction quality.

\textbf{Weight initialization for INRs. } Previous work has shown that smart initialization of INR weights allows for faster convergence. As shown in the \siren study~\cite{sitzmann2020implicit}, hyper-networks are proposed to capture a prior over the space of implicit functions, mapping a random code to the weights of a \siren model. Further, TransINR\cite{transinr} also shows Transformer-hypernetworks as powerful metalearners for INR weight initialization. MetaSDF \cite{sitzmann2020metasdf} and Light Field Networks (LFN) \cite{sitzmann2021light} use meta-learning-based initialization schemes to fit signed distance functions (SDFs) and light fields. Neural Compression algorithms~\cite{dupont2022coin++, strumpler2022implicit} use weights obtained from meta-learning optimization as a reference to store MLP weights, leading to better compression than naively storing MLP weights. Tancik et al.~\cite{tancik2021learned} propose meta-learning-based approaches such as Model-Agnostic Meta-Learning (MAML)\cite{finn17a} and Reptile\cite{nichol2018first} for coordinate based neural representations. While these meta-learning approaches yield powerful initialization, they often require long computation time (over 150K steps~\cite{tancik2021learned}) and ample numbers of training data, and are unstable to train~\cite{vettoruzzo2023advances}. Further, meta-learning initial modulations for an INR which are later optimized to fit data within few gradient updates has been shown to be an effective and scalable\cite{scalingfuncta} strategy for smoothly representing data(sets) as functa(sets)\cite{functa}. Contrary to our approach, Implicit Pattern Composers (IPC)\cite{ipc} proposes to keep the second layer of an INR instance-specific, while sharing the remaining layers and use a transformer hypernetwork to learn the modulations for the INR.

\textbf{Prior informed INRs. } Recent work has also explored embedding a learned prior in INRs for tasks such as audio compression, noise removal, and better CT reconstructions. Siamese Siren \cite{lanzendorfer2023siamese} uses a similar approach where they propose a compact siamese INR whose initial layers are shared followed by 2 siamese decoders. Since 2 randomly initialized decoders will yield slightly different reconstructions, this difference is leveraged for better noise estimation in audio signal. NERP \cite{shen2022nerp} learns an internal INR prior for medical imaging by first fitting high quality MRI or CT data. Weights of this learned INR are used as an initialization for reconstructing new MRI or CT undersampled data. While this paper shows a method to learn an implicit prior, their prior embedding is learned from a single MRI or CT scan of the same subject whereas our work explores learning a prior for INRs by constraining it to learn an underlying implicit representation across multiple different images. PINER \cite{song2023piner} introduces a test-time INR adaptation framework for sparse-view CT reconstruction with unknown noise.

%% file: content/3_Method/method.tex
We introduce \strainer. We first explain our motivation to share initial layers in an INR \Cref{sec:motivation}. In \Cref{sub_sec:shared_enc_train} we describe the training phase of \strainer where we learn transferrable features for INRs by sharing the initial layers of $N$ INRs being fit independently to $N$ images. \Cref{sub_sec:method_inference}, details how our captured basis is used to fit an unseen image. In subsequent sections, we seek to understand what our shared basis captures and how to expand it to other problems such as super resolution. For simplicity, we build upon the \siren \cite{sitzmann2020implicit} model as our base model.

\subsection{Why share the initial INR layers?}\label{sec:motivation}
A recent method called SplineCAM~\cite{humayun2023splinecam} provides a lens with which to visualize neural network partition geometries. SplineCAM interprets an INR as a function that progressively warps the input space and fits a given signal through layerwise affine transforms and non-linearities~\cite{humayun2023splinecam}. For continuous piecewise affine activation functions, we use an approximation to visualize (see \Cref{fig:partitions}) the deep network's partition geometry for different pre-activation level sets~\cite{humayun2024deep}. 

An INR fit to a signal highly adapts to the underlying structure of the data in a layer-wise fashion. Furthermore, by approximating the spatial position of the pre-activation zero level sets, we see that initial layers showcase a coarse, less-partitioned structure while deeper layers induce dense partitioning collocated with sharp changes in the image. Since natural signals tend to be similar in their lower frequencies, we hypothesize that initial layers of multiple INRs are better suited for \emph{transferability}. We therefore design \strainer to share the initial \textit{encoder} layers, effectively giving rise to an input space partition that can generalize well across different similar signals.

\subsection{Learning transferable features from $N$ images} \label{sub_sec:shared_enc_train}

Consider a \siren \cite{sitzmann2020implicit} model $h(p)$ with $L$ layers. Let $K$ out of $L$ layers correspond to an \textit{encoder} sub-network represented as $f_{\theta}$ The remaining layers correspond to the \textit{decoder} sub-network represented as $g_{\phi}$ as seen in \Cref{fig:coverfigure}(a). For given input coordinates $p$, we express the \siren model $h_{\phi, \theta}(p)$ as a composition ($\circ$) of our encoder-decoder sub-networks.
\begin{equation}
    h_{\phi, \theta}(p) = g_{\phi} \circ f_{\theta}(p) \text{ ,}
    \label{eq:network_subdivision}
\end{equation}
In a typical case, given the task of fitting $N$ signals, each of the $N$ signals is independently fit to an INR, thus not leveraging any similarity across these images. Since we want to learn a shared representation transferrable across multiple similar images, our method shares the encoder$f_{\theta}$ across all $N$ INRs while maintaining a set of individual signal-specific decoders $g_{\phi}^{1} \ \dots g_{\phi}^{N}$.Our overall architecture is shown in \Cref{fig:coverfigure}. We call this \strainers training phase - \Cref{fig:coverfigure}(a). We start with randomly initialized layers and optimize the weights to fit $N$ signals in parallel. 
For each signal $I_i(p)$, we use a $L_2$ loss between $I_i(p)$ and its corresponding learned estimate $h_{\phi, \theta}^{i}(p)$ and sum the loss over all the $N$ signals. Iteratively, we learn a set of weights $\Theta$ that minimizes the following objective:
\begin{equation}
    \Theta^* = \arg\min_{\Theta} \sum_{i=1}^{N} \mid \mid g_{\phi}^{i}\circ f_{\theta}(p) - I_i(p) \mid\mid_2^2 \text{ ,}
    \label{eq:train_n_decoders}
\end{equation}
where $\Theta = [\theta, \ \phi^1 \ \dots \phi^N]$ represents the full set of weights of the shared encoder ($\theta$) and the $N$ different decoders ($g_\phi^1 \dots g_\phi^N$) and $\Theta^*$ represents the resulting optimized weights. 

\subsection{Fitting an unseen signal with \strainerB} \label{sub_sec:method_inference}
After sufficient iterations during \strainers training phase, we get optimized encoder weights $f_{\theta^*}$ which corresponds to the rich shared representation learned over signals of the same category. To fit a novel signal $I_\psi(p)$ we initialize the \strainer model with the learned shared encoder weights $f_{\theta=\theta^*}$ and randomly initialize decoder $g_\phi^{\psi}$ weights to solve for:
\begin{equation}
    \phi^*,  \theta^* = \arg\min_{\phi, \theta} \mid\mid g_\phi^{\psi} \circ f_{\theta = \theta^*} (p)  - I_{\psi}(p) \mid\mid^2_2 \text{.}
    \label{eq:fit_test_general}
\end{equation}
$f_{\theta = \theta^*}$ serves as a learned initial encoder features. Our formulation is equivalent to a initial set of learned encoder features followed by a set of random projections. While fitting an unseen signal, we iteratively update all the weights of the \strainer model, similar to any INR. 

\begin{figure}[t!]
    \centering
    \includegraphics{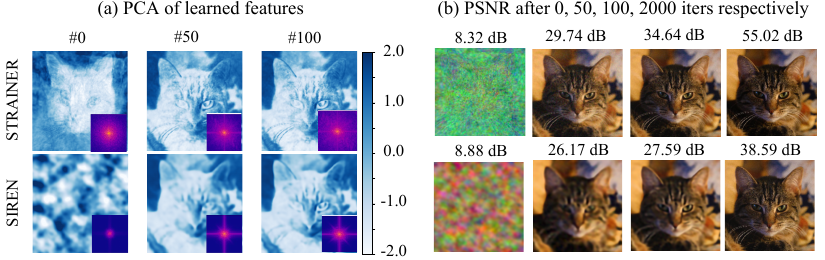}
    \caption{\textbf{Visualization of learned features in \strainerB and baseline \sirenB model}. We visualize (a) the first principal component of the learned encoder features for \strainer and corresponding layer for \siren. 
    At iteration 0, \strainers features already capture a low dimensional structure allowing it to quickly adapt to the cat image. High frequency detail emerges in \strainers learned features by iteration 50, whereas \siren is lacking at iteration 100. The inset showing the power spectrum of the reconstructed image further confirms that \strainer learns high frequency faster. We also show the (b) reconstructed images and remark that \strainer fits high frequencies faster.}
    \label{fig:pca_cat}
\end{figure}

\subsection{Learning an intermediate partition space in the shared encoder $f_{\theta^*}$}
During the training phase, explicitly sharing layers in \strainer allows us to learn a set of INR features which exhibits a common partition space shared across multiple images. Since deep networks perform layer wise subdivision of the input space, sharing the encoder enforces the layer to find the partition that can be further subdivided into multiple coarse partitions corresponding to the tasks being trained. 
In \Cref{fig:partitions}(a.ii), while pre-training an INR using the \strainer framework on \celeba dataset, we see emergence of a face-like structure captured in our \strainer encoder $f_{\theta^*}$. We expect our \strainer encoder weights $f_{\theta^*}$ to be used as transferrable features and be used as initialization for fitting unseen in-domain samples. 

In comparison, meta learning methods to learn initialization for INRs\cite{tancik2021learned} exhibit a partitioning of the input space that is closer to random. As seen in \Cref{fig:partitions}(a.i) there is a faint image structure captured by the the learned initialization. This is an indication that the initial subdivision of the input space, found by the meta learned pre-training layers, captures less of the domain specific information therefore is a worse initialization compared to \strainer. We further explain our findings in \Cref{sec:Discussions} and discuss \strainers learned features being more transferrable and lead to better quality reconstruction.

\begin{figure}[t!]
    \centering
    \includegraphics[width=1.0\textwidth]{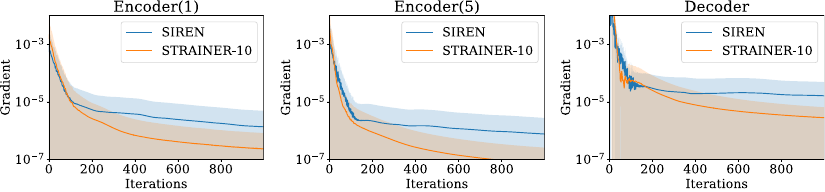}
    \caption{\strainerB\textbf{converges to low and high frequencies fast.} We plot the histogram of absolute gradients of layers 1,5 and last over 1000 iterations while fitting an unseen signal. At test time, \strainers initialization quickly learns low frequency, receiving large gradients update at the start in its initial layers and reaching convergence. The Decoder layer in \strainer also fits high frequency faster. Large gradients from corresponding \siren layers show it learning significant features as late as 1000 iterations.}
    \label{fig:gradient_iterations}
\end{figure}

%% file: content/4_ExperimentResults/Results.tex
In all experiments, we used the \citesiren MLP with $6$ layers and sinusoid nonlinearities.  We considered two versions of \strainer: (i) \strainer (1 decoder), where the encoder layers are initialized using our shared encoder trained on a single image, and (ii) \strainerten (10 Decoders), where the encoder layers are initialized using our shared encoder trained on 10 images. We considered the following baselines: (i) a vanilla \siren model with random uniform initialization~\cite{sitzmann2020implicit}, (ii) a fine-tuned \siren model initialized using the weights from another \siren fit to an image from the same domain, (iii) a \siren model initialized using \metalearning\cite{tancik2021learned}, (iv) transformer-based metalearning models such as TransINR\cite{transinr} and IPC\cite{ipc}. We ensured an equal number of learnable parameters (neurons) for all models. We normalized all images between (0-1), and input coordinates between (-1,1). We used the Adam optimizer with a learning rate of $10^{-4}$ for \strainers training and test-time evaluation, unless mentioned otherwise. Further implementation details are provided in Supplementary.

\subsection{Datasets}
We mainly used the CelebA-HQ \cite{celebahq}, Animal Faces-HQ (AFHQ) \cite{afhq}, and OASIS-MRI \cite{oasis1,oasis2} images for our experiments. We randomly divided CelebA-HQ into 10 train images and \TESTCASES test images. For AFHQ, we used only the cat data, and used ten images for training and \TESTCASESAFHQ images for testing. For OASIS-MRI, we used 10 of the (template-aligned) 2D raw MRI slices for training, and \TESTCASESMRI for testing. We also used Stanford Cars\cite{stanfordcars} and Flowers\cite{flowers} to further validate out of domain generalization and Kodak~\cite{kodak} true images for demonstrating high-resolution image fitting.

\subsection{Training \strainerBs shared encoder}
We first trained separate shared encoder layers of \strainer on \encoderimages train images from each dataset. We share five layers, and train a separate decoder for each training image. For each dataset, we trained the shared encoder for 5000 iterations until the model acheives PSNR $\approx 30dB$ for all training images. We use the resulting encoder parameters as initialization for test signals in the following experiments. For comparison, we also trained the \metalearning baseline using the implementation provided by Tancik et.al.\cite{tancik2021learned} with 5000 outer loop iterations. We also use the implementation provided by IPC\cite{ipc} as our baselines for TransINR\cite{transinr} and IPC\cite{ipc} and train them with 14,000 images from \celeba. We report a comparison of number of training images and parameters, gradient updates, and learning time in \Cref{tab:init_runtime}.

\subsection{Image fitting (in-domain)}
We first evaluated \strainer on the task of in-domain image fitting. We cropped and resized all images to $178 \times 178$ and ran test-time optimization on all models for $2000$ steps.

At test-time, both \strainer and \strainerten use only 1 decoder, resulting in the same number of parameters as a \siren INR. 
\Cref{tab:celeba_results} shows average image metrics for in-domain image fitting reported with 1 std. deviation. Instead of naively fine tuning using another INR, \strainers design of sharing initial layers allows for learning highly effective features which transfer well across images in the same domain, resulting in high quality reconstruction across \celeba and AFHQ and comparable to \metalearning for OASIS-MRI images. \Cref{tab:baseline_w_transinr_ipc}(\celeba, ID) also shows that \strainer initialization results in better quality reconstruction, when optimized at test-time, compared to more recent transformer-based INR approaches such as TransINR and IPC as well.

\begin{table}[t!]
    \centering
    \caption{\textbf{In-domain image fitting evaluation.} \strainers learned features yield powerful initialization at test-time resulting in high quality in-domain image fitting}   
    \label{tab:celeba_results}
    \begin{adjustbox}{width=\textwidth,center}
    \begin{tabular}{lllllll}
    \toprule
    \multicolumn{1}{c}{} & \multicolumn{2}{c}{CelebA-HQ} & \multicolumn{2}{c}{AFHQ}  & \multicolumn{2}{c}{OASIS-MRI}\\
    \cmidrule(lr){2-3}\cmidrule(lr){4-5}\cmidrule(lr){6-7}
      Method  & PSNR$\uparrow$ & SSIM$\uparrow$  & PSNR$\uparrow$ & SSIM$\uparrow$ & PSNR$\uparrow$ & SSIM$\uparrow$\\
      \midrule
      \siren &  44.91 $\pm$ 2.13 & 0.991 $\pm$ 0.007 &  45.11 $\pm$ 3.13  & 0.991 $\pm$ 0.005   &  53.03 $\pm$ 1.72 & 0.999 $\pm$ 0.0002\\
      \siren fine-tuned & 51.11 $\pm$ 3.16 & 0.997 $\pm$  0.013  &  53.07 $\pm$ 3.47 & 0.997  $\pm$ 0.001  &  58.86 $\pm$ 4.12 & 0.999 $\pm$ 0.0012\\
      \metalearning & 53.08 $\pm$ 3.36 & 0.994 $\pm$ 0.053 &  53.27 $\pm$  2.52& 0.996 $\pm$ 0.044 &  67.02 $\pm$ 2.27 & 0.999 $\pm$ $0.0000$ \\
      \strainer(1 decoder) & 50.34 $\pm$ 2.81 & 0.997 $\pm$ 0.001 &  51.27 $\pm$ 2.94 & 0.997 $\pm$  0.001  &  57.76 $\pm$ 2.19 & 0.999 $\pm$ 0.0001\\
      \strainerten & 57.80 $\pm$ 3.46 & 0.999 $\pm$ 0.001 &  58.06 $\pm$ 3.75 & 0.999 $\pm$ 0.001  &  62.80 $\pm$ 3.17 & 0.999 $\pm$ 0.0003\\
      \bottomrule
    \end{tabular}
    \end{adjustbox}
\end{table}

\begin{figure}[t!]
    \centering
    \includegraphics{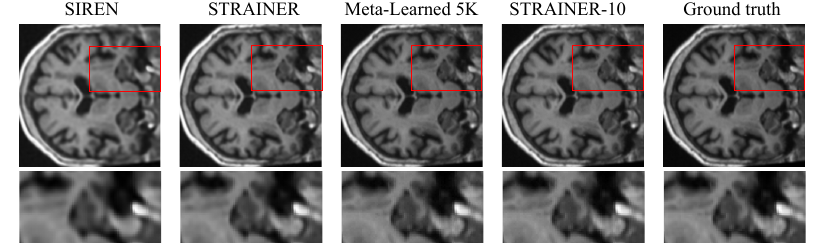}
    \caption{\textbf{Fitting MRI images from OASIS-MRI dataset.} At just 100 iterations, \strainer is able to represent medical images with high quality. \strainers initialization allows for fast recovery for sparse and delicate structures, showing applicability in low-resource medical domains as well.}
    \label{fig:mri_structure}
\end{figure}

\begin{table}[t!]
    \centering
    \caption{\textbf{Out of domain image fitting evaluation, when trained on \celeba and tested on AFHQ and OASIS-MRI.} \strainers learned features are a surprisingly good prior for fitting images out of its training domain.}  
    \label{tab:outdomain_fit}
    \begin{adjustbox}{width=0.9\textwidth,center}
    \begin{tabular}{lllll}
    \toprule
    \multicolumn{1}{c}{} & \multicolumn{2}{c}{AFHQ}  & \multicolumn{2}{c}{OASIS-MRI}\\
    \cmidrule(lr){2-3}\cmidrule(lr){4-5}
      Method  & PSNR$\uparrow$ & SSIM$\uparrow$ & PSNR$\uparrow$ & SSIM$\uparrow$\\
      \midrule
      \metalearning  &  52.40 $\pm$ 4.21 & 0.991 $\pm$ 0.077 &  65.06 $\pm$ 1.04 & 0.999 $\pm$ 0.00001\\
    
      \strainerten  &  57.46 $\pm$ 3.39& 0.999 $\pm$ 0.0003   &  72.21 $\pm$ 8.73 & 0.999 $\pm$ 0.0001\\
      \strainerten(Gray) & -- & -- & 74.61 $\pm$ 9.96 & 0.999 $\pm$ 0.0003\\
      \bottomrule
    \end{tabular}
    \end{adjustbox}
\end{table}

\begin{table}[h!]
    \centering
    \caption{\textbf{Baseline evaluation for image-fitting for in-domain(ID) and out-of-domain(OD) data.} \strainer learns more transferable features resulting in better performance across the board. Models trained on \celeba unless mentioned otherwise. TTO = Test time optimization.}
    \label{tab:baseline_w_transinr_ipc}
    \begin{adjustbox}{width=0.85\textwidth,center}
    \begin{tabular}{llll}
    \toprule
        \multicolumn{1}{c}{} & \multicolumn{1}{c}{CelebA-HQ (ID)} & \multicolumn{1}{c}{AFHQ (OOD)} & \multicolumn{1}{c}{OASIS MRI (OOD)}\\
        Method & PSNR$\uparrow$  & PSNR$\uparrow$ & PSNR$\uparrow$ \\
        \midrule
        \metalearning & 53.08  &  52.40 & 55.86 \\
        \transinrdefault  & 31.59  & 28.63 &  31.97  \\
        \transinr  & 51.86  &  49.01 &  55.45 \\
        \ipcreludefault  &  33.27   & 29.96  & 33.96 \\
        \ipcrelu  & 49.72 &  47.19 &  51.35  \\
        \midrule
        \strainerten  & \textbf{57.80}  & \textbf{57.46}  & 59.50 \\
        \strainerten ( trained on Flowers\cite{flowers}) & - & 56.98 & 58.52\\
        \strainerten ( trained on StanfordCars\cite{stanfordcars}) & - & 56.88 & \textbf{59.66}\\
        \bottomrule
    \end{tabular}
    \end{adjustbox}
\end{table}

\begin{table}[h!]
    \centering
    \caption{\textbf{Out-of-domain image fitting on Kodak Images \cite{kodak}}. \strainer (trained on \celeba) allows better convergence comparable to high capacity \siren models as indicated by PSNR metric.}
    \label{tab:eval_kodak_id}
    \begin{adjustbox}{width=1.0\textwidth,center}
    \begin{tabular}{lllllllllll}
    \toprule
        \multicolumn{1}{c}{} & \multicolumn{1}{c}{} & \multicolumn{3}{c}{Parrot} & \multicolumn{3}{c}{Airplane}  & \multicolumn{3}{c}{Statue}\\
    \cmidrule(lr){3-5}\cmidrule(lr){6-8}\cmidrule(lr){9-11}
    Method & Width  & PSNR$\uparrow$ & SSIM$\uparrow$ & LPIPS$\downarrow$ & PSNR$\uparrow$ & SSIM$\uparrow$ & LPIPS$\downarrow$ & PSNR$\uparrow$ & SSIM$\uparrow$ & LPIPS$\downarrow$\\
       \midrule
          \siren & 256 & 36.77  & 0.94 & 0.13  & 31.89 & 0.87 & 0.19 & 34.68 & 0.94 & 0.093 \\ 
          \strainerten & 256 & 39.55 & 0.96 & 0.087 & 35.03 & 0.92 & 0.09 & 37.84 & 0.96 & 0.037\\
          \siren & 512 & 40.18 & 0.96 & 0.11 & 34.23 &  0.90 & 0.14& 38.80 & 0.97 & 0.051 \\
          \strainerten & 512 &  44.38 & 0.97 & 0.021 & 38.96 & 0.96 & 0.023& 43.92 & 0.98 & 0.008\\
      \bottomrule
    \end{tabular}
    \end{adjustbox}
\end{table}

\subsection{Image fitting (out-of-domain)}
To test out-of-domain transferability of learned \strainer features, we used \strainerten's encoder trained on \celeba as initialization for fitting images from AFHQ (cats) and OASIS-MRI datasets (see \Cref{tab:outdomain_fit}). Since OASIS-MRI are single channel images, we trained \metalearning and \strainerten(GRAY) on the green channel only of \celeba images. To our surprise, we see \strainerten and \strainerten(GRAY) clearly outperform not only \metalearning, but also \strainerten (in-domain). To further validate out of domain performance of \strainer, we train \strainerten's shared encoder on  simply 10 images from Flowers\cite{flowers} and Stanford Cars\cite{stanfordcars} datasets which have different spatial distribution of color and high frequencies than AFHQ and OASIS-MRI.  For fair comparison, all models in \Cref{tab:baseline_w_transinr_ipc}(OOD) were fit with 3-channel RGB or concatenated gray images in case of OASIS-MRI. As shown in \Cref{tab:baseline_w_transinr_ipc}(OOD), \strainerten provides superior out of domain performance for AFHQ trained on \celeba, followed by Flowers and Stanford Cars. For OASIS-MRI, we see \strainerten having best performance when trained with StanfordCars. This result suggests that \strainer is capable of capturing transferable features that generalize well across natural images.

\strainer also fits high resolution Kodak\cite{kodak} images well and is comparable to \siren networks with twice the network width. 

\subsection{Inverse problems: super-resolution and denoising}
\strainer provides a simple way to encode data-driven priors, which can accelerate convergence on inverse problems such as super-resolution and denoising. 
We sampled $100$ images from \celeba at $178\times178$ and added $2dB$ of Poisson random noise. We report mean values of PSNR achieved by \strainer and \siren models along with the iterations required to achieve the values. For super-resolution, we demonstrate results on one image from DIV2K\cite{div2k1,div2kw}, downscaled to $256\times 256$ for a low resolution input. We used the formulation shown in \Cref{eq:L2_loss}, with $\mathcal{A}$ set to a $4\times$ downsampling operation. To embed a prior relevant for clean images, we trained the shared encoder of \strainer with high quality images of resolution same as the latent recovered image. At test time, we fit the \strainer model to the corrupted image, following \Cref{eq:L2_loss} and recovered the latent image during the iteration. We report \strainers ability to recover latent images fast as well as with high quality in \Cref{tab:inv_problems}

\begin{table}[t!]
    \centering
    \caption{\textbf{Training time and compute complexity.} We train all the methods for $5000$ steps. \strainer instantly learns a powerful initialization with minimal data and significantly fewer gradient updates.}
    \label{tab:init_runtime}
    \begin{adjustbox}{width=1.0\textwidth,center}   
    \begin{tabular}{lllll}
        \toprule
          Method & \# training images & \# learnable params  &  Gradient updates / iteration & Time (Nvidia A100)\\
          \midrule
             \siren & N/A & 264,707 & N/A & N/A \\
             \strainer (1 decoder) & 1 & 264,707 & 264,707  & 11.84s\\
             \strainerten (10 decoders) & 10 &  271,646 & 271,646 & 24.54s\\
             \metalearning & 10 & 264,707& 794,121 ($\approx 3\times$more) &  1432.3s = 23.8 min\\
             TransINR\cite{transinr} & 14,000 & $\approx 40M$ & $\approx 40M$ & $\approx 1$ day\\
             IPC\cite{ipc} w TTO & 14,000 & $\approx 40M$ & $\approx 40M$ & $\approx 1$ day\\
        \bottomrule
    \end{tabular}
    \end{adjustbox}
\end{table}


\begin{table}[t!]
    \centering
    \label{tab:inv_problems}
    \caption{\textbf{\strainer accelerates recovery of latent images in inverse problems.} \strainer captures an implicit prior over the training data allowing it to recover a clean latent image of comparable quality $3\times$ faster making it useful for inverse problems.}
    \begin{adjustbox}{width=\textwidth, center}
    \begin{tabular}{lllllll}
        \toprule
        \multicolumn{1}{c}{} & \multicolumn{2}{c}{Super Resolution (Fast)} & \multicolumn{2}{c}{Super-Resolution (HQ)} & \multicolumn{2}{c}{Denoising}\\
        \cmidrule(lr){2-3}\cmidrule(lr){4-5}\cmidrule(lr){6-7}
          Method & PSNR & \# iterations & PSNR & \# iterations & PSNR & \# iterations \\
          \midrule
          \siren & 32.10  & 3329 & 32.10 & 3329 & 26.75 $\pm$ 1.67 & 203 $\pm$ 66\\
          \strainer-10 & 31.56 & 1102 ($\approx 3\times faster$)  & 32.43 & 3045 &26.41 $\pm$ 1.39 & 76 $\pm$ 27 \\
          \bottomrule
    \end{tabular}
    \end{adjustbox}
\end{table}

%% file: content/7_Discussions/Discussions.tex
\begin{figure}[t!]
    \centering
    \includegraphics[width=\textwidth]{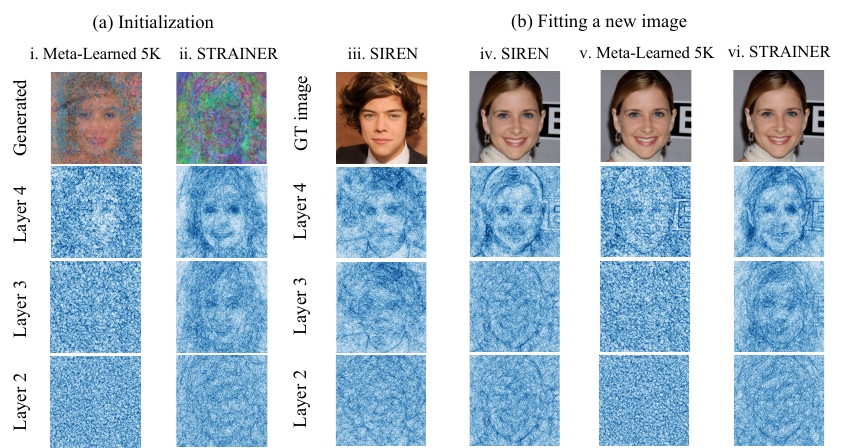}
    \caption{\textbf{Visualizing density of partitions in input space of learned models.} We use the method introduced in \cite{humayun2024deep} to approximate the input space partition of the INR. 
    We present the input space partitions for layers 2,3,4 across (a) \metalearning and \strainer initialization and (b) at test time optimization. \strainer learns an input space partitioning which is more attuned to the prior of the dataset, compared to meta learned which is comparatively more random. We also observe that \siren(iii) learns an input space partitioning highly specific to the image leading to inefficient transferability for fitting a new image (iv) with significantly different underlying partitioned input space
    This explains the better in-domain performance of \strainer compared to \metalearning, as the shallower layers after pre-training provide a better input space subdivision to the deeper layers to further subdivide.
    }
    \label{fig:partitions}
    \vspace{-10pt}
\end{figure}

Results in Table~\ref{tab:celeba_results}, \ref{tab:baseline_w_transinr_ipc} demonstrate that \strainer can learn a set of transferable features across an image distribution to precisely fit unseen signals at test time. \strainerten clearly achieves the best reconstruction quality in terms of PSNR and SSIM on \celeba and AFHQ datasets, and is comparable with \metalearning on OASIS-MRI images. \strainerten also fits images fast and achieves highest reconstruction quality than all baselines as shown in \Cref{fig:psnr_plot_all}.
Comparing \strainer(1 decoder) with a fine-tuned \siren, it seems that the representation learned on one image is not sufficiently powerful. However, as little as 10 images result in a rich and transferable set of INR features allowing \strainerten to achieve $\approx$7-10dB higher reconstruction quality than \siren and \siren fine-tuned. 

As seen in \Cref{tab:outdomain_fit}, \ref{tab:baseline_w_transinr_ipc}(OOD) \strainer also performs well on out-of-domain tasks, which is quite surprising. 

\strainers transferable representations are capable of recovering small and delicate structures as early as 100 iterations as shown in \Cref{fig:mri_structure} and do not let the scale of features from the training phase affect its reconstruction ability. Another interesting finding is that \strainerten achieves far better generalization for OASIS-MRI (\Cref{tab:outdomain_fit}) when pretrained on \celeba. Further, \strainer generalizes well to out-of-domain high-resolution images, as demonstrated by our experiments of training \strainer on \celeba and testing on the Kodak data (see \Cref{tab:eval_kodak_id}).

\strainer is fast and cheap to run. \Cref{tab:init_runtime} summarizes the time for learning the initialization for a $6$ layered MLP INR for \strainer, \metalearning and transformer-based methods such as TransINR and IPC. At $5000$ iterations, \strainer learns a transferable representation in just $24.54$ seconds.  \metalearning, in comparison, uses MAML\cite{finn17a} which is far more computationally intensive and results in $20\times$ slower runtime when exact number of gradient updates are matched. Further, \strainer's training setup is an elegant deviation from recent methods such as TransINR and IPC, requiring large datasets and complex training routines.

\subsection{Limitations} 
\label{sec:Limitations}
Due to the encoder layers of \strainer being tuned on data and the later layers being randomly initialized, we have observed occasional instability when fitting to a test signal in the form of PSNR ``drops.'' However, we observe that \strainer usually quickly recovers, and the speedup provided by \strainer outweighs this issue. While our work demonstrates that INR parameters may be transferred across signals, it is not fully clear what features are being transferred, how they change for different image distributions, and how they compare to the transfer learning of CNNs and Transformers. Further work is needed to characterize these. 

\subsection{Further analysis of \strainerB}\label{sec:strainer_frequencies}
To further understand how \strainers initialized encoder enables fast learning of signals at test time, we explored the evolution of \strainers hidden features over iterations in \Cref{fig:pca_cat}. 
In \Cref{fig:pca_cat}(a), we visualize the first principal component of learned INR features of the \strainer encoder and corresponding hidden layer for \siren across iterations and observe that \strainer captures high frequencies faster than \sirenn. This is corroborated by the power spectrum inset plots of the reconstructed images. We also visualize a histogram of gradient updates in \Cref{fig:gradient_iterations}, and observe that \strainer receives large gradients in its encoder layers early on during training, 
suggesting that the encoder rapidly learns of low-frequency details. 

Next, we visualize the input space partitions induced by \strainer and the adaptability of \strainers initialization for fitting new signals. We use the local complexity(LC) measure proposed by Humayun et.al.\cite{humayun2024deep} to approximate different pre-activation level sets of the INR neurons. For ReLU networks, the zero level sets correspond to the spatial location of the non-linearities of the network. For periodic activations, there can be multiple non-linearities affecting the input domain. In \Cref{fig:partitions} we present the zero level sets of the network, and in Supplementary we provide the $\pm \pi/2$ shifted level sets. Darker regions in the figure indicate high LC, i.e., higher local non-linearity. \Cref{fig:partitions} also presents partitions for the baseline models.

\siren models tend to overfit, with partitions strongly adapting to image details. Since the sensitivity to higher frequencies is mapped to specific input partitions, when finetuning with \siren, the network has to unlearn partitions of the pretrained image resulting in sub optimal reconstruction quality. When comparing \metalearning with \strainer, we see that \strainer learns an input space partitioning more attuned to the prior of the dataset, compared to \metalearning which is comparatively more random. While both partitions imply learning high-frequency details, \strainers partitions are better adapted to facial geometry, justifying its better in-domain performance.

%% file: content/6_BroaderImpact/BroaderImpact.tex
\strainer introduces how to learn transferable features for INRs resulting in faster convergence and higher reconstruction quality. We show with little data, we can learn powerful features as initialization for INRs to fit signals at test-time.  Our method allows the use of INRs to become ubiquitous in data-hungry areas such as patient specific medical imaging, personalized speech and video recordings, as well as real-time domains such as video streaming and robotics. However, our method is for training INRs to represent signals in general, which can adopted regardless of underlying positive or negative intent.

%% file: content/Acknowledgements/Ack.tex
This work is partially supported by NIH DeepDOF R01DE032051-01, OneDegree CNS-1956297, IARPA WRIVA 140D0423C0076 and NSF grants CCF1911094, IIS-1838177, and IIS-1730574; ONR grants N00014- 18-12571, N00014-20-1-2534, and MURI N00014-20-1-2787; AFOSR grant FA9550-22-1-0060; and a Vannevar Bush Faculty Fellowship, ONR grant N00014-18-1-2047.

%% file: content/Supplementary/supp.tex
\section*{Suplementary Material / Appendix}

In our work, we increase the representation capacity of the INR by leveraging the similarity across natural images (of a given class). Since each layer of an INR MLP is an affine transformation followed by a non linearity, we interpret the INR as a function that progressively warps the input coordinate space to fit the given signal, in our case the signal being an image. Similar images when independently fit to their respective INRs capture similar low-frequency detail such as shape, geometry, etc. whereas high-frequency information such as edges and texture are unique to each INR. We propose that these low-frequency features from the initial layers of a learned INR are highly transferable and can be used as a basis and initialization while fitting an unseen signal. To that end, we introduce a novel method of learning our basis by sharing a set of initial layers across INRs fitting their respective images. 

Our implementation can be found on \href{https://colab.research.google.com/drive/1fBZAwqE8C_lrRPAe-hQZJTWrMJuAKtG2?usp=sharing}{\footnote{https://colab.research.google.com/drive/1fBZAwqE8C\_lrRPAe-hQZJTWrMJuAKtG2?usp=sharing}{Google Colab}}.

\section*{Understanding the effect of sharing encoder layers}
\begin{figure}[h!]
    \centering
    \includegraphics[width=0.6\textwidth]{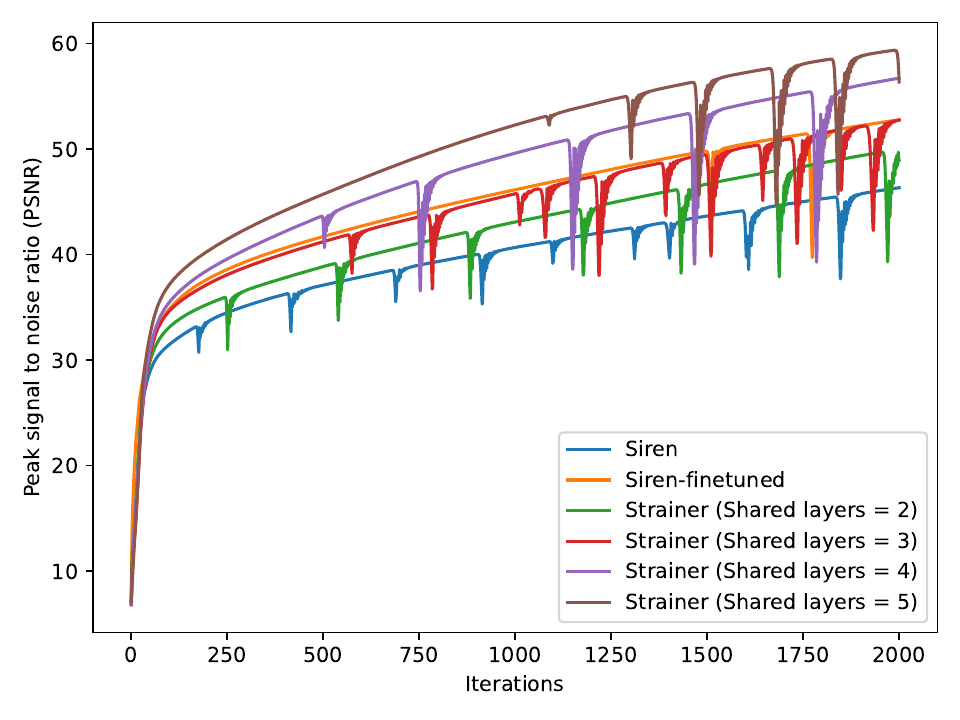}
    \caption{Sharing different number of layers in \strainers encoder. We see that by increasing the number of shared layers, \strainers ability to recover the signal also improves.}
    \label{fig:num_shared_layers_plot}
\end{figure}

We further investigate how the number of initial layers shared affects the quality of reconstructed image. We start by sharing $K=1$ layer as the encoder, and $N-K$ layers in each decoder and vary $K$ from 1 to $N-1$.  $K=N$ is equivalent to simply fine tuning the INR based on all weights from a fellow model. We tabulate our results for image quality (PSNR) in a fixed runtime of 1000 iterations. We find that sharing all but the last layer results in the most effective capturing of our shared basis leading to higher reconstruction quality as seen in \cref{fig:num_shared_layers_plot}. This also suggests that the last decoder of the INR is mainly responsible for very localized features. Further our work motivates further interest to sutdy the nature of the decoder layers itself. 

We show the effect of sharing layers and resulting reconstruction quality. We use a 5 layered Siren model for this experiment. We fit a vanilla Siren model to an image and report its PSNR in \cref{fig:num_shared_layers_plot}. Further, we train our shared encoder by sharing $K=1$ layers and so on , until we share $K=N-1$ layers.

We see that the reconstruction quality progressively increases by sharing layers.

\begin{figure}[h!]
    \centering
    \includegraphics[width=\textwidth]{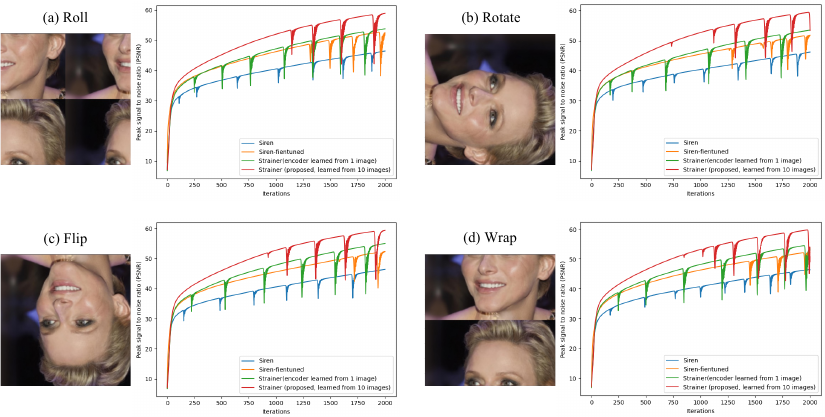}
    \caption{Fitting \strainer on shifted, rotated and flipped versions of a face image. We see that despite the transformations done on a face image: (a) roll (wrapping the image both vertically and horizontally), (b) rotate, (c) flip vertically, and (d) wrap vertically, \strainer fits equally well on all of them.}
    \label{fig:orientation}
\end{figure}

\section*{Reporting std. deviation \strainer for image fitting on \celeba}
\begin{figure}[h]
    \centering
    \includegraphics[width=0.5\linewidth]{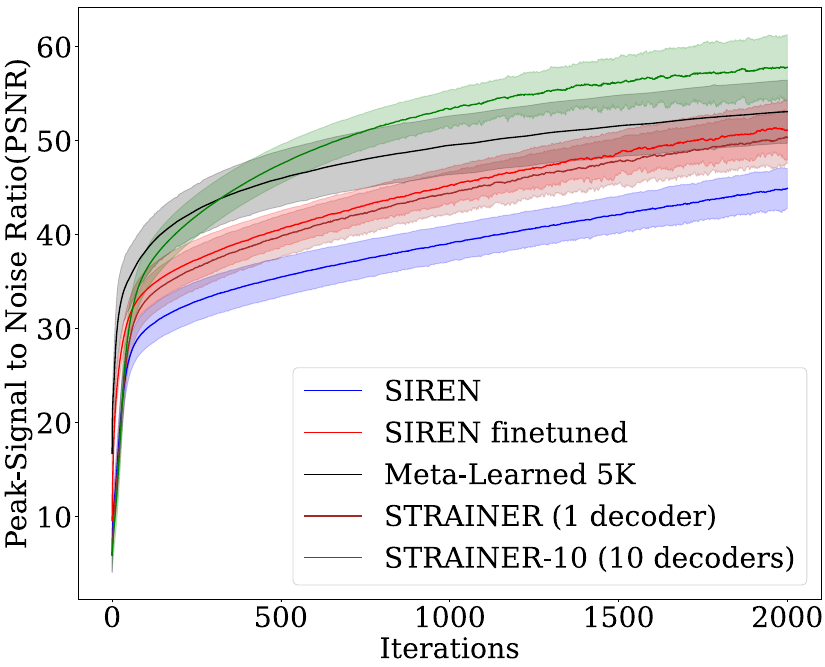}
    \caption{\strainer learns fast. We show a limited baseline comparison of \strainer with \siren, \siren-finetuned and \metalearning methods for the task of image-fitting on \celeba dataset and note that \strainer achieves superior reconstruction quality.}
    \label{fig:supp_psnr_std}
\end{figure}

We also report the PSNR within 1 std. deviation while comparing \strainer-10 with \siren, \siren-finetuned, \strainer(1-decoder), and \metalearning in \Cref{fig:supp_psnr_std}.

\section*{Effect of orientation of the input image}
We wanted to further assess whether the INR is overfitting to one particular \textit{aligned} face arrangement. To further test this, we take a test image and apply various augmentations such as flip, rotate, and \textit{roll} and study how \strainer fits it (see \Cref{fig:orientation}).  We find that the initialization learned by strainer is invariant, at test time, to the input signals orientation and can successfully capture the high frequency details fast.

\section*{Measuring time for pretraining \strainerB and \metalearning}
Our implementation is written in PyTorch\cite{pytorch} whereas \metalearning implemented by Tancik et.al\cite{tancik2021learned} is a JAX implementation. For measuring runtime, we use the python's \textit{time} package and very conservatively time the step where the forward pass and gradient updates occur in both methods. Further, we run the code on an Nvidia A100 GPU and report the time after averaging 3 such runs for each method. There may be system level differences, however, to the best of our knowledge and observation, our timing estimates if not accurate are atleast indicative of the speedup provided by \strainer.

\section*{Training details for Kodak high resolution images}

To further demonstrate that \strainers can be adapted to high resolution images, we evaluated our method on high quality Kodak\cite{kodak} images with resolution $512 \times 768$ (see \Cref{tab:eval_kodak_id,tab:eval_kodak_supp_od}). We present the reconstruction quality attained by \strainer-10, \siren model, and a \siren model initialized using \metalearning, with widths of $256,512$. For this experiment, we train our \strainer encoder using \celeba Images which are resized to the same resolution to Kodak images. Further, we follow all steps as previously described for test-image evaluation of Kodak images. Here is another results from the Kodak high resolution images experiment.

\begin{table}[h!]
    \centering
    \caption{\strainer allow better convergence comparable to high capacity Siren models, and meta-learned initializations, as indicated by PSNR metric. Tested on high quality Kodak Images. ID = In domain, OD= Out of domain.}
    \label{tab:eval_kodak_supp_od}
    \begin{adjustbox}{width=1.0\textwidth,center}
    \begin{tabular}{lllllllllllll}
    \toprule
         \multicolumn{1}{c}{} & \multicolumn{3}{c}{Parrot (OD)} & \multicolumn{3}{c}{Airplane (OD)}  & \multicolumn{3}{c}{Statue (OD)} & \multicolumn{3}{c}{Painted Face(ID)}\\
    \cmidrule(lr){1-1}\cmidrule(lr){2-4}\cmidrule(lr){5-7}\cmidrule(lr){8-10} \cmidrule(lr){11-13}
    {Width=256}  & PSNR$\uparrow$ & SSIM$\uparrow$ & LPIPS$\downarrow$ & PSNR$\uparrow$ & SSIM$\uparrow$ & LPIPS$\downarrow$ & PSNR$\uparrow$ & SSIM$\uparrow$ & LPIPS$\downarrow$ & PSNR$\uparrow$ & SSIM$\uparrow$ & LPIPS$\downarrow$\\
       \midrule
       \siren &  36.77  & 0.94 & 0.13  & 31.89 & \textbf{0.87} & 0.19 & 34.68 & 0.94 & 0.093 &  32.03 & 0.85 & 0.26 \\
       \strainerten & \textbf{39.55} & \textbf{0.96} & 0.087 & \textbf{35.03} & 0.92 & \textbf{0.09} & \textbf{37.84} & \textbf{0.96} & \textbf{0.037} &  \textbf{35.15} & \textbf{0.92} & \textbf{0.11}\\
       \metalearning & 37.07 & 0.94 & \textbf{0.06} &  33.92 & 0.89 & 0.12 & 34.32 & 0.93 & 0.07 &  32.96 & 0.89 & 0.11 \\
        \midrule
        {Width=512}  & PSNR$\uparrow$ & SSIM$\uparrow$ & LPIPS$\downarrow$ & PSNR$\uparrow$ & SSIM$\uparrow$ & LPIPS$\downarrow$ & PSNR$\uparrow$ & SSIM$\uparrow$ & LPIPS$\downarrow$ & PSNR$\uparrow$ & SSIM$\uparrow$ & LPIPS$\downarrow$\\
         \cmidrule(lr){1-1}\cmidrule(lr){2-4}\cmidrule(lr){5-7}\cmidrule(lr){8-10} \cmidrule(lr){11-13}
         \siren  & 40.18 & 0.96 & 0.11 & 34.23 &  0.90 & 0.14& 38.80 & 0.97 & 0.051  &  34.45 & 0.90 & 0.17 \\
          \strainerten  &  \textbf{44.38} & \textbf{0.97} & 0.021 & 38.96 & 0.96 & 0.023& \textbf{43.92} & \textbf{0.98} & \textbf{0.008} & \textbf{41.37} & \textbf{0.97} &  \textbf{0.006} \\
          \metalearning & 41.60 & 0.97 & \textbf{0.02} & \textbf{39.33} & 0.96 & \textbf{0.02} & {39.18} & 0.97 & 0.02 & 37.90 & 0.96 & 0.03 \\
      \bottomrule
    \end{tabular}
    \end{adjustbox}
\end{table}

\section*{Results for Inverse problems - Super Resolution}
\begin{figure}[h!]
    \centering
    \includegraphics{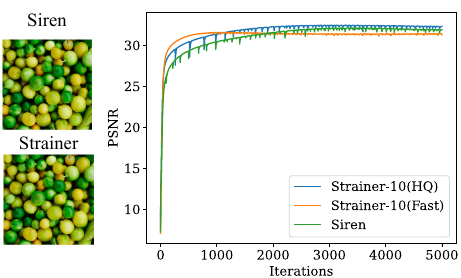}
    \caption{Super Resolution using \strainer. We show the reconstructed results (a) on the left using \siren and \strainer. We also plot (b) the trajectory of PSNR with iterations. \strainerten(Fast) achieves comparable PSNR to \siren in approximately a third of the runtime.}
    \label{fig:sr_supp_strainer}
\end{figure}

\begin{figure}[h!]
    \centering
    \includegraphics[width=0.5\textwidth]{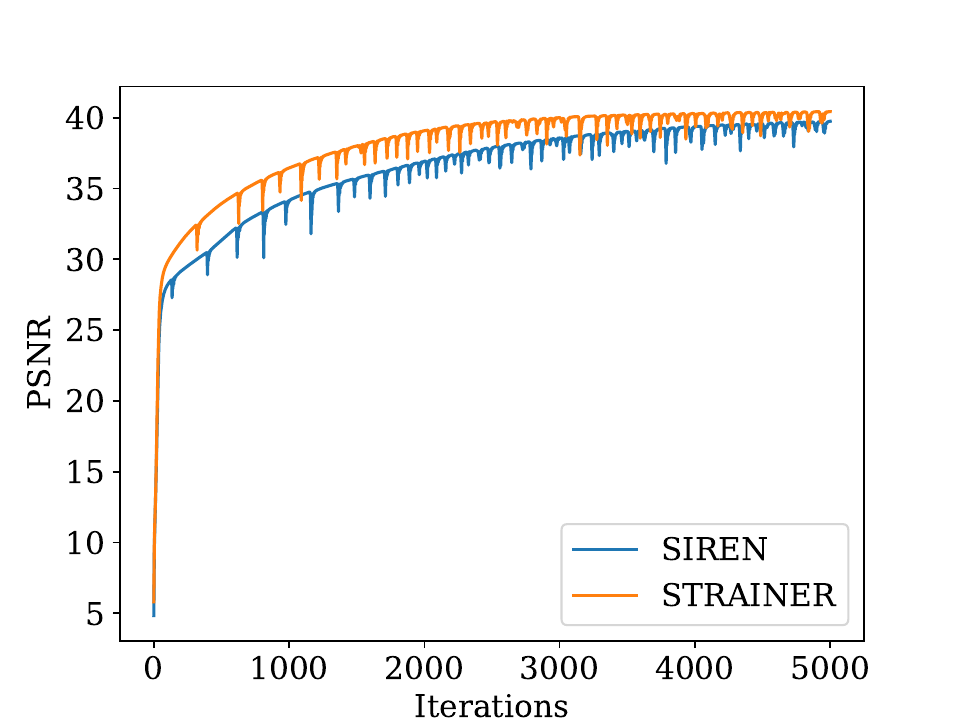}
    \caption{In domain $4\times$ super resolution using \strainer. We see that \strainer allows for faster convergence for in-domain super resolution making it useful especially for low time budgets. Max value achieved by \strainer: $40.43dB$ while \siren achieves $39.75dB$. Within 500 iterations \strainer achieves $>30dB$ PSNR}
    \label{fig:in_domain_sr}
\end{figure}

We discuss how \strainer provides a useful prior for inverse problems such as super resolution. For the results reported in \cref{tab:inv_problems}, we attach supplementary plots as shown in \cref{fig:sr_supp_strainer}. \strainerten(Fast) is a \strainerten model with $5$ shared encoder layers out of $6$ total layers. \strainerten(HQ) is a high quality \strainer model with $3$ shared encoder layers. Unlike forward fitting, more degree of randomness in the decoder helps recover better detail for inverse problems. We also showcase the effectiveness of \strainer for in domain super resolution shown in \cref{fig:in_domain_sr}.

\section*{\strainerB for Occupancy fitting} 
\begin{figure}[h!]
    \centering
    \includegraphics[width=0.9\linewidth]{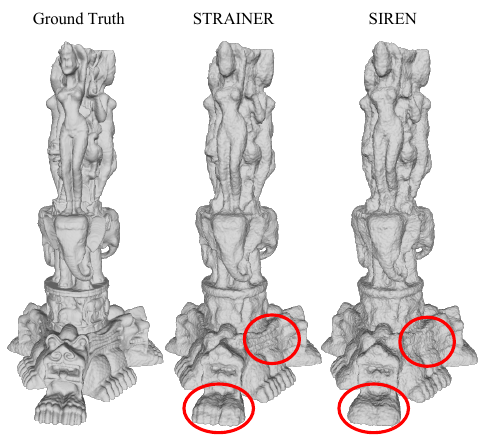}
    \caption{We use ten shapes from the chair category of ShapeNet\cite{shapenet2015} to train STRAINER , and use that initialization to fit a much more complex volume (the Thai statue\cite{thaistatue}). We compare the intermediate outputs for both STRAINER and SIREN for 150 iterations to highlight STRAINER ’s ability to learn ridges and high frequency information faster.
}
    \label{fig:supp_occupancy}
\end{figure}

STRAINER is a general purpose transfer learning framework which can be used to initialize INRs for regressing 3D data like occupancy maps, radiance fields or video. To demonstrate the effectiveness of STRAINER on 3D data, we have performed the following OOD generalization experiment. We pre-train STRAINER on 10 randomly selected ‘Chair’ objects from the ShapeNet\cite{shapenet2015} dataset. At test time, we fit the ‘Thai Statue’ 3D object\cite{thaistatue}. STRAINER achieves a 12.3 relative improvement in IOU compared to random initialization for a SIREN architecture – in 150 iterations STRAINER-10 obtains an IOU of 0.91 compared to an IOU of 0.81 without STRAINER-10 initialization. We present visualizations of the reconstructed Thai Statue in \Cref{fig:supp_occupancy}. Upon qualitative evaluation, we see that STRAINER-10 is able to capture ridges and edges better and faster than compared to SIREN. 

\newpage
\section*{Offsetting Pre-activations}

\begin{figure}[h!]
    \centering
    \includegraphics{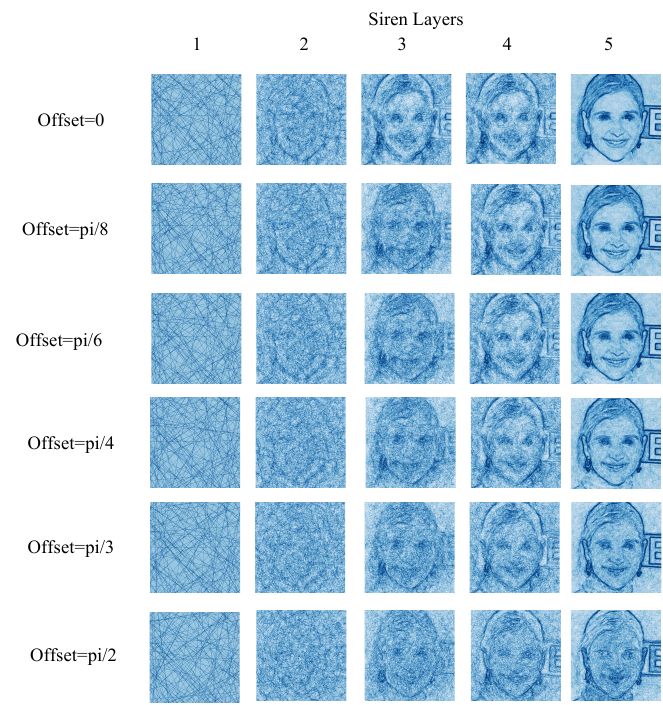}
    \caption{Siren offset}
    \label{fig:enter-label}
\end{figure}
\begin{figure}[h!]
    \centering
    \includegraphics{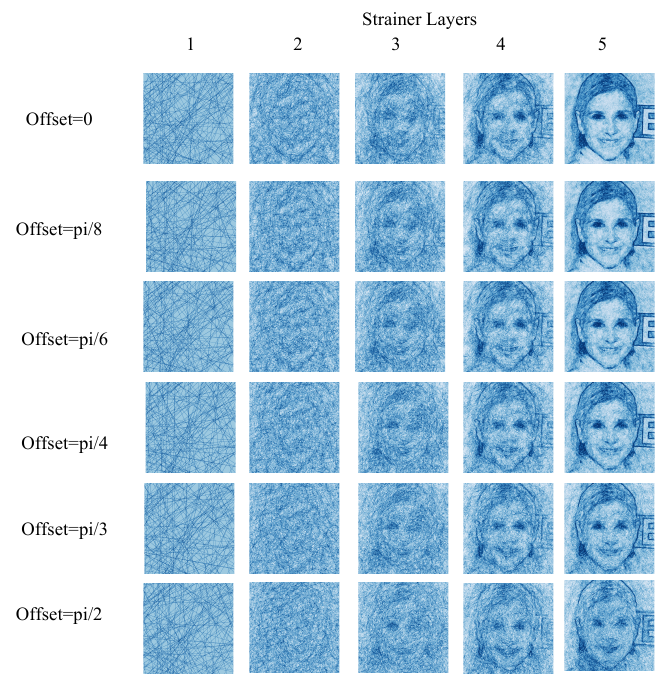}
    \caption{Strainer offset}
    \label{fig:enter-label}
\end{figure}
\begin{figure}[h!]
    \centering
    \includegraphics{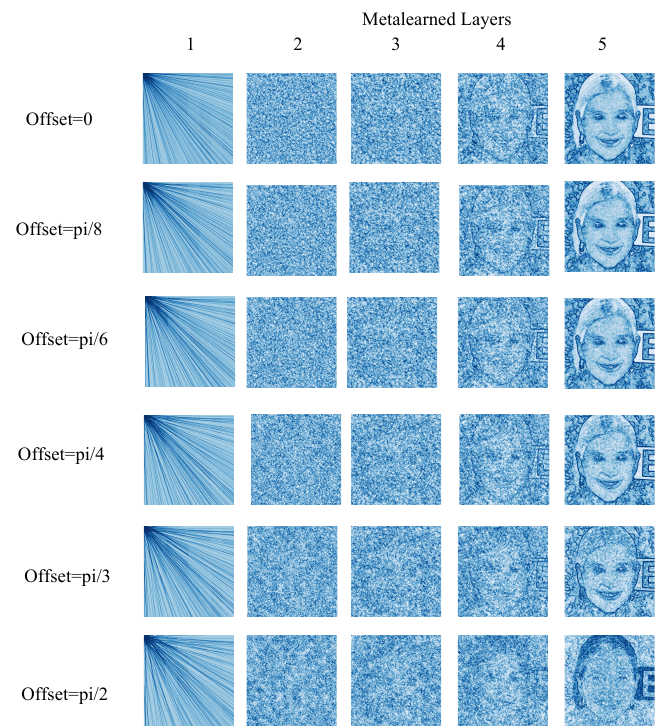}
    \caption{\metalearning offset}
    \label{fig:enter-label}
\end{figure}